\pdfoutput=1
\documentclass{article}
\usepackage{spconf,amsmath,graphicx}
\usepackage{textcomp}
\usepackage{stfloats}
\usepackage{url}
\usepackage{verbatim}
\usepackage{subfig}
\usepackage{multirow}
\usepackage{graphicx}
\usepackage{cite}
\usepackage{booktabs}
\usepackage{float}
\usepackage[T1]{fontenc}
\usepackage[utf8]{inputenc}
\usepackage{babel}
\usepackage{booktabs}
\usepackage{setspace}
\usepackage{amssymb}
\usepackage{bbding}
\usepackage{pifont}
\usepackage{wasysym}
\usepackage{utfsym}
\usepackage{fontawesome}
\usepackage{makecell}
\usepackage{amsmath,amsfonts}
\usepackage{algorithmic}
\usepackage{algorithm}
\usepackage{array}
\usepackage{threeparttable}    
\usepackage{setspace}
\usepackage[colorlinks,linkcolor=red]{hyperref}

\title{A general framework for rotation invariant point cloud analysis}
\name{Shuqing Luo, Wei Gao\thanks{This work is supported by Natural Science Foundation of China (62271013, 62031013), Shenzhen Fundamental Research Program (GXWD20201231165807007-20200806163656003), Shenzhen Science and Technology Program (JCYJ20230807120808017), and Sponsored by CAAI-MindSpore Open Fund, developed on OpenI Community (CAAIXSJLJJ-2023-MindSpore07). Wei Gao is the corresponding author.}}
\address{School of Electronic and Computer Engineering, Peking University Shenzhen Graduate School \\ $\texttt{luoshuqing@stu.pku.edu.cn, gaowei262@pku.edu.cn}$}
\begin{document}
\begin{sloppypar}
%
\maketitle
\begin{abstract}
We propose a general method for deep learning based point cloud analysis, which is invariant to rotation on the inputs. Classical methods are vulnerable to rotation, as they usually take aligned point clouds as input. Principle Component Analysis (PCA) is a practical approach to achieve rotation invariance. However, there are still some gaps between theory and application. In this work, we present a thorough study on designing rotation invariant algorithms for point cloud analysis. We first formulate it as a permutation invariant problem, then propose a general framework which can be combined with any backbones. Our method is beneficial for further research such as 3D pre-training and multi-modal learning. Experiments show that our method has considerable or better performance compared to state-of-the-art approaches on common benchmarks. Code is available at \url{https://github.com/luoshuqing2001/RI\_framework}.
\end{abstract}
\begin{keywords}
Point cloud learning, rotation invariant, PCA, general framework
\end{keywords}
\section{Introduction}
Deep learning has been successfully applied to point cloud recognition tasks, achieving outstanding performance\cite{qi2017pointnet, qi2017pointnet++, wang2019dynamic, li2018pointcnn, qian2022pointnext, li2018so, riegler2017octnet}. However, mainstream algorithms typically process aligned point clouds as input, neglecting rotated cases. Existing studies have demonstrated that randomly rotated point cloud can lead to terrible performance using traditional algorithms.

So far, numerous rotation robust or invariant methods have been proposed  for point cloud learning models. These can be broadly categorized as follows:
\begin{itemize}
    \item Rotation robust methods. These approaches aim to align the input into a suitable pose\cite{fang2020rotpredictor}. As this is conducted by prediction, the alignment may be invalid.
    \item Rotation invariant methods based on handcrafted features or sorted feature map. These work try to design rotation invariant geometric features such as distance between two points, angle between two edges or dihedral angle, and feed them into neural networks as input\cite{li2021rotation, zhang2019rotation, zhang2022riconv++, xiao2021triangle, zhao2022lgr-net, rao2019spherical}. Besides, some work also arrange the feature map into certain order, which would not be affected by rotation transformation\cite{xu2021sgmnet, kim2020rotation-nips}.
    \item Rotation invariant methods based on Principle Component Analysis (PCA). When processed by PCA, the input point cloud would deduce two matrices (consist of eigenvalues and eigenvectors, respectively). The multiplication result between point cloud and one matrix can be proved to be rotation invariant. These computation results can also be called canonical poses, which can be used for the neural networks\cite{li2021closer, yu2020deeppca}. 
\end{itemize}

PCA based method is most competitive and promising among them due to its interpretability, provability and scalability. The interpretability benefits from the transparency of PCA, which can be viewed as additional rotation to the input point cloud. The provability is because the whole computing process can be rigorously proved to be rotaion invariant mathematically. The scalability is because the core procedure of point cloud recognition is still the classical backbones, which can facilitate further studies such as 3D pre-training and multi-modal learning.

However, the computation result of PCA is uncertain, which is ignored by most of the previous work\cite{li2021closer}. The size of solution space is finite, but the order of it is uncertain. 

In this work, we first model the PCA based rotation invariant point cloud learning as a permutation invariant problem. Then we design a general framework, which can be combined with classical point cloud learning backbones. We summarize our contributions as follows:
\begin{itemize}
    \item We take a thorough study on the PCA process for point cloud, and point out some attributes of canonical poses, which are significant for rotation invariant learning.
    \item We propose a framework for point cloud learning, which can be combined with common backbone networks and is provably invariant to rotation.
    \item We propose a data augmentation method in our framework, which substantially enhances the performance.
\end{itemize}

\section{Formulation for canonical poses}
In this section, we formulate the PCA process for point cloud. For a given input $\mathbf{P}\in\mathbb{R}^{N\times3}$ ($N$ denotes the number of points) and rotation transformation $\mathbf{R}\in\mathbb{R}^{3\times3}$, the rotated point cloud is $\mathbf{P}\mathbf{R}$. The PCA process for $\textbf{P}$ can be formulated as
\begin{equation}\label{PCA}
    \frac{1}{N}\sum\limits_{i=1}^N(\mathbf{P}_i-\Bar{\mathbf{P}})^{\top}(\mathbf{P}_i-\Bar{\mathbf{P}})=\mathbf{E}\Lambda\mathbf{E}^{\top},
\end{equation}
where $\Bar{\mathbf{P}}\in\mathbb{R}^3$ is the center of $\mathbf{P}$, $\Lambda$ is the diagonal matrix composed of three eigenvalues, i.e, $\Lambda=\text{diag}(\lambda_1, \lambda_2, \lambda_3)$, and $\mathbf{E}$ is composed of eigenvectors, i.e, $\mathbf{E}=(e_1, e_2, e_3)$. It is easy to prove that, for rotated $\mathbf{P}'=\mathbf{P\cdot R}$, the corresponding matrix $\mathbf{E}'=\mathbf{R}^{\top}\mathbf{E}$. If $\mathbf{E}$ does not change with rotation, then we can get the canonical pose $P_{\mathrm{cano}}$ for both $\mathbf{P}$ and $\mathbf{P}'$ as
\begin{align}\label{canonical}
\begin{split}
    \mathbf{P}_{\mathrm{cano}}&=\mathbf{P}\mathbf{E} \\
    &=(\mathbf{P}\mathbf{R})(\mathbf{R}^{\top}\mathbf{E}),
\end{split}
\end{align}
which shows the rotation invariance of canonical pose. 

For point cloud with normal vectors denoted as $\mathbf{P}^{''}\in\mathbb{R}^{N\times6}$, $\mathbf{E}$ is still got from $xyz$ coordinates. Specifically, $\mathbf{P}^{''}_{\mathrm{cano}}=\mathbf{P}^{''}\mathbf{E}^{''}$, in which
\begin{equation}
\mathbf{E}^{''}=
\begin{pmatrix}
    \mathbf{E} &  \\
     & \mathbf{E}
\end{pmatrix}.
\end{equation}

\subsection{Canonical Pose Space}
We take a deeper look at the details. Define $\mathbf{M}=\frac{1}{N}\sum\limits_{i=1}^N(\mathbf{P}_i-\Bar{\mathbf{P}})^{\top}(\mathbf{P}_i-\Bar{\mathbf{P}})$, the PCA process on input point cloud can be decomposed as
\begin{equation}\label{decomposition}
    \mathbf{M} e_i = \lambda_i e_i,
\end{equation}
for $i\in\{1,2,3\}$. The sign of $e_i$ does not have an influence on the equation, thus there are $2^3$ cases for $\mathbf{E}$ when the order of eigenvalues is fixed. Furthermore, if we take the order of eigenvalues into consideration, there are $3!$ cases for $\mathbf{E}$ when the sign of each eigenvector is fixed. Therefore, the whole canonical pose space can be formulated as 
\begin{equation}\label{48-cases}
    6\text{ (Order Cases)}\times 8\text{ (Sign Cases)} = 48
\end{equation}
cases.

If we directly take the whole space as input for the learning model, it is obvious that the redundant computation is too high. So we need to reduce the size of canonical pose space to improve computational efficiency while ensuring provable rotation invariance. 

We take $E_0$ as an example, assuming that $\det{(E_0)}=1$ and the corresponding $\Lambda_0$ is in descending sort. For the total $48$ cases, we first keep the $8$ matrices with eigenvalues in descending order while removing the $40$ matrices in other order. Secondly, we keep the rest $4$ matrices with determinant value $1$ while removing the $4$ matrices with determinant value $-1$. In this way, the size of canonical pose space can be reduced from $48$ to $4$. The final canonical space with $4$ poses can be used as the direct input of point cloud recognition models, and it is irreducible.

\subsection{Attributes of Canonical Pose space}
Now we present two attributes of canonical pose space, which have not been explored by previous work. They are uncertainty in PCA process and operational closure, respectively.

\subsubsection{Uncertainty in PCA process}
We consider an input point cloud at object level with random rotation. The PCA process program (e.g, \texttt{torch.linalg.eigh} in PyTorch and \texttt{numpy.linalg.eigh} in Numpy) can only get one instance of matrix $\mathbf{E}$ within total 4 cases, and this solution result would vary along with different rotation matrices. Therefore, the order of the $4$ canonical poses is uncertain given the same input point cloud under different rotation. If we adopt PCA to achieve rotation invariance in the learning algorithm, then we can only model it as a permutation invariant problem.

\subsubsection{Operational closure}
As the order of canonical pose space is uncertain, we can only model it as a set. Now we prove that the canonical pose space with size $4$ is consistent in all cases. Assuming that $E_0=(e_1, e_2, e_3)$ satisfying $\det(E_0)=1$, and we also define $4$ matrices as $T_1=\text{diag}(1,1,1),\ T_2=\text{diag}(-1,-1,1),\ T_3=\text{diag}(-1,1,-1)$ and $T_4=\text{diag}(1,-1,-1)$. The canonical pose space can be formulated as $S=\{E_i\|\i=1,2,3,4\}$, in which $E_i=E_0T_i$. As mentioned above, for any matrix $\mathbf{E}$ derives from PCA program satisfying both $\det(\mathbf{E})=1$ and eigenvalues in descending order, there must be $\mathbf{E}\in S$. Besides, we can also obtain set $S$ from any instance $\mathbf{E}\in S$ by implementing matrix multiplication with $T=\{T_i\|\i=1,2,3,4\}$, respectively. This can be formulated as, for any $\mathbf{E}\in S$, there is $S=\{ET_i\|\i=1,2,3,4\}$. It is easy to prove this attribute of operational closure, as the number of total states is limited. 


\section{Our method}
As the size of canonical pose space cannot be less than 4, and the order of it is uncertain, we can only compute 4 feature maps individually, and fuse them at different level. In this way, we can turn any hierarchical point cloud recognition backbone networks into a rotation invariant version, so that to realize generality and scalability. 

\begin{figure}[h]
\centering
\includegraphics[scale=0.06]{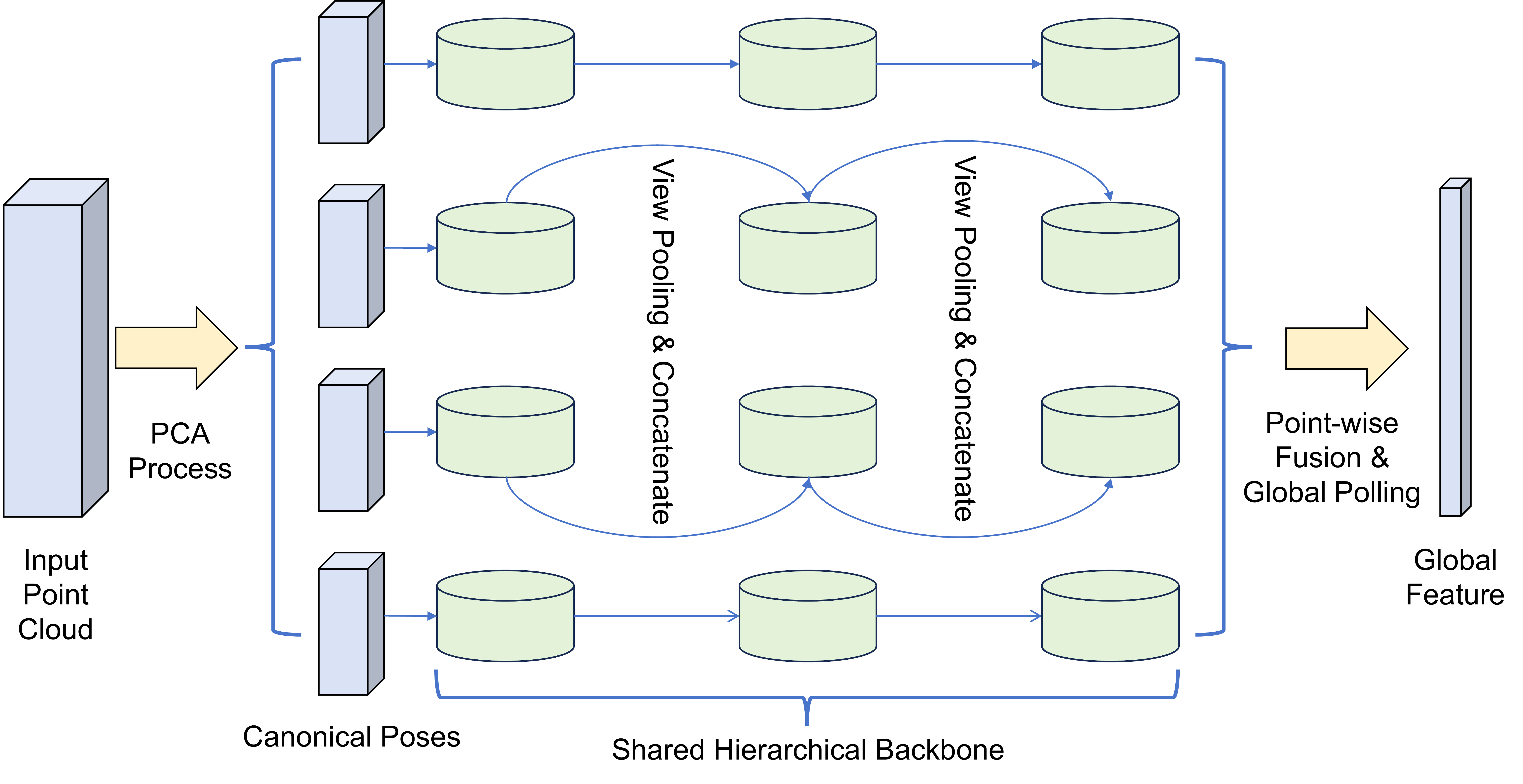}
\caption{Our general framework for rotation invariant point cloud analysis. PCA transforms the input point cloud into $4$ canonical poses, while shared hierarchical backbone extracts local feature for each view and each point. Feature fusion is implemented on modules at the same stage.}
\label{fig:framework}
\end{figure}

In this section, we propose two feature fusion methods at different level, which is view pooling and point-wise fusion. They are simple but efficient in point cloud recognition tasks. The detailed network architecture is shown in Fig.\ref{fig:framework}, in which the four canonical poses share the same hierarchical backbone, and in this figure it is split into $N$ modules. Feature fusion is implemented within modules at same stage.

\subsection{View pooling among modules at same stage}
View pooling is implemented within the same stage. This is inspired by PointNet++ and other hierarchical backbones, as max pooling is implemented at each stage within some range of point sets. In our algorithm, the feature map with length $\mathbf{M}$ after module $i$ for canonical pose $j$ \big($i\in\{1,2,\cdots,N\}$ and $j\in\{1,2,\cdots,4\}$\big) is denoted as $f_{ij}$. View pooling is defined as 
\begin{equation}
    v_i[k] = \max_{j\in\{1,2,\cdots,4\}} f_{ij}[k], \ \forall{k\in\{1,2,\cdots, \mathbf{M}\}}.
\end{equation}

Then $v_i$ is concatenated with $f_{ij}$ as the input of module at next stage. This can make an module obtain the feature of $4$ canonical poses before it, so that to better fuse the useful information. 

\subsection{Point-wise fusion before global pooling}
At last stage, we can get point-wise and view-wise feature maps. We denote it as $\big\{P_{ij}\|\i\in\{1,2,\cdots,N_0\},j\in\{1,2,\cdots,4\}\big\}$, in which $N_0$ is the number of points at last stage. This can be less than input, as down sampling may be included in the backbone. A shared multi-layer perception module (MLP) is required. We denote it as $\mathcal{F}$.

First, we fuse the $4$ feature of each point using $\mathcal{F}$. $\mathcal{F}$ predicts weight for each view. Specifically, $\mathcal{F}$ is implemented on $\{P_{ij}\}$ for each $i$ and $j$:
\begin{equation}
    w_{ij} = \mathcal{F}(P_{ij}), \ \forall i\in\{1,2,\cdots,N_0\}, j\in\{1,2,\cdots,4\}.
\end{equation}

$\{w_{ij}\}$ is then normalized channel-wise by Softmax. We denote the length of feature map as $\mathbf{M}'$:
\begin{equation}
    w_{ij}[k] = \frac{\exp{(w_{ij}[k])}}{\sum_{j=1}^4\exp{(w_{ij}[k]})}, \ \forall k\in\{1,2,\cdots, \mathbf{M}'\}.
\end{equation}

Then the feature map of 4 views with each point is fused by channel-wise dot product:
\begin{equation}
    P_i[k] = \sum_{j=1}^4 w_{ij}[k]\cdot P_{ij}[k], \ \forall k\in\{1,2,\cdots, \mathbf{M}'\}.
\end{equation}

Finally, global pooling can be implemented on this fused $\{P_i\}$ for downstream tasks like classification or segmentation. 

As we can see, our method adds relatively few parameters compared to the original backbone network. Inference can be done in parallel, as the 4 branches play the same role. Thus the time cost would not increase significantly, although there would be approximately quadruple memory required.

\subsection{Data Augmentation}
To further improve the performance of our method as well as enhance the robustness of recognition, we propose a data augmentation method named \emph{random scaling}. Let's rethink the PCA process, which can be summarized as rotating an input point cloud with random pose to a canonical one adaptively. The canonical pose is determined by the shape of raw input, and scaling transformation can change it to a large extent. We denote the scaling matrix as $\textbf{S}=\text{diag}(a,b,c)$, input point cloud as $\textbf{P}\in\mathbb{R}^{N\times3}$. The scaled point cloud is $\textbf{P}'=\textbf{PS}$. Correspondingly, the canonical pose is
\begin{equation}
    \textbf{P}_{\mathrm{cano}}'=\textbf{PS}'\textbf{E},
\end{equation}
in which 
\begin{equation}
    \textbf{S}'=\text{diag}\left(\frac{a^2}{(abc)^\frac{1}{3}}, \frac{b^2}{(abc)^\frac{1}{3}}, \frac{c^2}{(abc)^\frac{1}{3}}\right).
\end{equation}

Therefore, we can change $\textbf{S}$ to get different canonical poses so as to expand the training dataset. The algorithm is shown in Algorithm.\ref{alg:alg1}. This procedure is the same for point cloud with normal vectors.

\begin{algorithm}[H]
\begin{footnotesize}
\begin{algorithmic}[1]
\REQUIRE $p\in\mathbb{R}^{N\times3}, \mathbf{M}\in\mathbb{N}$
\ENSURE $p^{\prime}\in\mathbb{R}^{N\times3}$
\STATE \textbf{Generate Random Vectors:}
\STATE \hspace{0.5cm} $\text{Randomly Generate }A\in\mathbb{R}^{3}\text{, }A[i]\in(1, \mathbf{M})$
\STATE \hspace{0.5cm} $\text{Randomly Generate }B\in\mathbb{R}^{3}\text{, }B[i]\in\{0, 1\}$
\STATE \textbf{Compute Transformation Matrix:}
\STATE \hspace{0.5cm} $\text{Initialize }C=\text{diag}(1,1,1)$
\STATE \hspace{0.5cm} $\textbf{for}\text{ i in }\{0, 1, 2\}\textbf{ do}$
\STATE \hspace{1.0cm} $\textbf{if }B[i] == 0\textbf{ then}$
\STATE \hspace{1.5cm} $C[i][i]=A[i]$
\STATE \hspace{1.0cm} $\textbf{else}:$
\STATE \hspace{1.5cm} $C[i][i]=1/A[i]$
\STATE \hspace{1.0cm} $\textbf{end if}$
\STATE \hspace{0.5cm} $\textbf{end for}$
\STATE \textbf{Matrix Multiplication:}
\STATE \hspace{0.5cm} $\textbf{return }$ $p'=p\cdot C$
\end{algorithmic}
\caption{Random Scaling Augmentation}
\label{alg:alg1}
\end{footnotesize}
\end{algorithm}

\section{Experiments}
\subsection{Datasets and Evaluation Metrics}
In this paper, experiments are implemented on two tasks, \i.e, point cloud classification on ModelNet40 dataset and point cloud part segmentation on ShapeNet part segmentation dataset. Performance is tested in 3 modes following previous benchmarks\cite{li2021rotation, sun2019srinet, yu2020deeppca}, which are
\begin{itemize}
    \item $z/z$: Both training and testing datasets are rotated around $z$-axis.
    \item $z/$\text{SO(3)}: Training dataset is rotated around the $z$-axis and testing dataset is randomly rotated.
    \item \text{SO(3)/SO(3)}: Both training and testing datasets are randomly rotated.
\end{itemize}

\subsection{Experiment Results}

The classification and part segmentation results are shown in Table.\ref{modelnet40} and Table.\ref{shapenet}. Without normal, our method has a competitive result on ModelNet40 and \textbf{SOTA} result on ShapeNet part segmentation. Performance comparison between origin algorithm and its rotation invariant version is shown in Table.\ref{comparison}. For aligned dataset, the performance of our framework is close to the raw backbone, while for the rotated data, our framework has obvious advantage.

\begin{table}[h]
    \centering
    \caption{Classification accuracy (\%) on the ModelNet40 dataset.}
    \resizebox{1.0\columnwidth}{!}{
    \begin{tabular}{c|cccccc}
        \hline
        \text{} & \text{Method} & \text{Input} & $z/z$ & $z$\text{/SO(3)} & \text{SO(3)/SO(3)} & \text{Drop of Acc}\\
        \hline
        \multirow{4}*{Rotation sensitive} 
        & \text{PointNet} & \text{xyz}$(1024\times3)$ & 88.5 & 16.4 & 70.5 & 54.1 \\
        & \text{PointNet++} & \text{xyz+normal}$(5000\times6)$ & 91.9 & 18.4 & 74.7 & 56.3 \\
        & \text{SO-Net} & \text{xyz+normal}$(5000\times6)$ & \textbf{93.4} & 19.6 & 78.1 & 58.5 \\
        & \text{DGCNN} & \text{xyz}$(1024\times3)$ & 91.2 & 16.2 & 75.3 & 59.1 \\
        \hline
        \multirow{4}*{Rotation robust} 
        & \text{Spherical CNN} & \text{voxel} & 88.9 & 76.9 & 86.9 & 10.0 \\
        & $a^3$\text{SCNN} & \text{voxel} & 89.6 & 87.9 & 88.7 & 0.8 \\
        & \text{SFCNN} & \text{xyz+normal}$(1024\times6)$ &  92.3 & 85.3 & 91.0 & 5.7 \\
        & \text{RotPredictor} & \text{xyz+normal}$(5000\times6)$ & 93.9 & 88.6 & 89.9 & 1.3 \\
        \hline
        \multirow{9}*{Rotation invariant} 
        & \text{Triangle-Net} & \text{xyz}$(1024\times3)$ & \text{-} & \text{-} & 86.7 & \text{-} \\
        & \text{RIConv} & \text{xyz}$(1024\times3)$ & 86.5 & 86.4 & 86.4 & 0.0 \\
        & \text{RI-GCN} & \text{xyz+normal}$(1024\times6)$ &  91.0 & 91.0 & 91.0 & 0.0 \\
        & \text{SRI-Net} & \text{xyz}$(1024\times3)$ & 87.0 & 87.0 & 87.0 & 0.0 \\
        & \text{Li }\textit{et al.} & \text{xyz}$(1024\times3)$ & 89.4 & 89.4 & 89.3 & 0.1 \\
        & \text{SGMNet} & \text{xyz}$(1024\times3)$ & 90.0 & 90.0 & 90.0 & 0.0 \\
        & \text{LGR-Net} & \text{xyz+normal}$(1024\times6)$ & 90.9 & 90.9 & 91.1 & 0.2 \\
        & \text{RIConv++} & \text{xyz+normal}$(1024\times6)$ & 91.3 & \textbf{91.3} & \textbf{91.3} & 0.0 \\
        & \text{Ours+PointNet++} & \text{xyz}$(1024\times3)$ & 90.4 & 90.4 & 90.4 & 0.0 \\
        \hline
    \end{tabular}
    }
    \label{modelnet40}
\end{table}

\begin{table}[h]
    \centering
    \caption{Mean IoU (\%) over all instances on ShapeNet part segmentation dataset.}
    \resizebox{1.0\columnwidth}{!}{
    \begin{tabular}{c|ccccc}
        \hline
        \text{} & \text{Method} & \text{Input} & $z$\text{/SO(3)} & \text{SO(3)/SO(3)} & \text{Drop of mIoU}\\
        \hline
        \multirow{4}*{Rotation sensitive} 
        & \text{PointCNN} & \text{xyz}$(1024\times3)$ & 34.7 & 71.4 & 36.7 \\
        & \text{DGCNN} & \text{xyz}$(1024\times3)$ & 37.4 & 73.3 & 35.9 \\
        & \text{PointNet} & \text{xyz}$(1024\times3)$ & 37.8 & 74.4 & 36.6 \\
        & \text{PointNet++} & \text{xyz+normal}$(5000\times6)$ & 48.2 & 76.7 & 28.5 \\
        \hline
        \multirow{10}*{Rotation robust \& invariant} 
        & \text{Triangle-Net} & \text{xyz}$(1024\times3)$ &  \text{-} & 72.5 & \text{-} \\
        & \text{RIConv} & \text{xyz}$(1024\times3)$ & 75.3 & 75.5 & 0.2 \\
        & \text{RI-GCN} & \text{xyz}$(1024\times3)$ & 77.2 & 77.3 & 0.1 \\
        & \text{RotPredictor} & \text{xyz}$(2048\times3)$ & 82.1 & 77.6 & 4.5 \\
        & \text{SRI-Net} & \text{xyz}$(1024\times3)$ & 80.0 & 80.0 & 0.0 \\
        & \text{Li }\textit{et al.} & \text{xyz}$(1024\times3)$ & 82.2 & 82.5 & 0.3 \\
        & \text{SGMNet} & \text{xyz}$(1024\times3)$ & 79.3 & 79.3 & 0 \\
        & \text{LGR-Net} & \text{xyz+normal}$(1024\times6)$ & \text{-} & 82.8 & \text{-} \\
        & \text{RIConv++} & \text{xyz+normal}$(1024\times6)$ & 80.5 & 80.5 & 0 \\
        & \text{Ours+DGCNN} & \text{xyz}$(1024\times3)$ & \textbf{84.5} & \textbf{84.5} & 0.0 \\
        \hline
    \end{tabular}
    }
    \label{shapenet}
\end{table}

\begin{table}[h]
    \centering
    \caption{Performance comparison between origin algorithm and the rotation invariant version.}
    \resizebox{0.9\columnwidth}{!}{
    \begin{tabular}{c|c|c|cccc}
        \hline
        \makecell{Task \& Metric} & \text{Backbone} & \text{Type} & \text{aligned} & $z/z$ & $z$\text{/SO(3)} & \text{SO(3)/SO(3)} \\
        \hline
        \multirow{2}*{\makecell{Classofication \\ (Accuracy, \%)}} & \multirow{2}*{PointNet++} 
        & \text{w/o ours} & $\textbf{91.9}$ & $\textbf{91.9}$ & $18.4$ & $74.7$ \\
        & & \text{w/ ours} & $90.3$ & $90.3$ & $\textbf{90.3}$ & $\textbf{90.3}$ \\
        \hline
        \multirow{2}*{\makecell{Part Segmentation \\ (Mean IoU, \%)}} & \multirow{2}*{DGCNN} 
        & \text{w/o ours} & $\textbf{85.2}$ & $82.3$ & $37.4$ & $73.3$ \\
        & & \text{w/ ours} & $84.5$ & $\textbf{84.5}$ & $\textbf{84.5}$ & $\textbf{84.5}$ \\
        \hline
    \end{tabular}
    }
    \label{comparison}
\end{table}

\subsection{Attribute Comparison}

Comparison on different attributes with current algorithms is shown in Table.\ref{tab:comparison_on_attributes}. We can see that our method is the only one that owns three advantages simultaneously while also has a quite high performance. 

\begin{table}[h]     
\centering                
\caption{Comparison on Different Attributes with Current Rotation Robust \& Invariant Methods}  
\vspace{0.15cm}      
\label{tab:comparison_on_attributes}     
\begin{threeparttable}
\resizebox{0.9\columnwidth}{!}{
\begin{tabular}{c|c|c|c|c}
\hline
\text{} & \text{Method} & \text{Scalability}\tnote{1} & \text{Rotation Invariance} & \text{No need for Handcrafted Feature} \\
\hline
\multirow{4}*{Rotation robust} 
& \text{Spherical CNN} &  &  & \CheckmarkBold \\
& $a^3$\text{SCNN} &  &  &  \\
& \text{SFCNN} &  &  &  \\
& \text{RotPredictor} & \CheckmarkBold &  &  \\
\hline
\multirow{9}*{Rotation invariant} 
& \text{Triangle-Net} &  & \CheckmarkBold  &  \\
& \text{RIConv} &  & \CheckmarkBold &  \\
& \text{RI-GCN} &  & \CheckmarkBold & \CheckmarkBold \\
& \text{SRI-Net} &  & \CheckmarkBold &  \\
& \text{Li }\textit{et al.} &  & \CheckmarkBold &  \\
& \text{SGMNet} & \CheckmarkBold & \CheckmarkBold &  \\
& \text{LGR-Net} &  & \CheckmarkBold &  \\
& \text{RIConv++} &  & \CheckmarkBold &  \\
& \text{Ours} & \CheckmarkBold & \CheckmarkBold & \CheckmarkBold \\
\hline
\end{tabular}
}
\begin{tablenotes}    
\footnotesize               
\item[1] Scalability denotes whether the method can be combined with \\ common point cloud learning backbones. 
\end{tablenotes} 
\end{threeparttable}
\end{table}

\subsection{Ablation Study}

Ablation study within $4$ fusion schemes at last stage is shown in Table.\ref{ablation}. We can see that our proposed method has the best performance compared with other common schemes. This is because point-wise fusion before global pooling can obtain as much useful information as possible.

\begin{table}[h]
    \centering
    \caption{Ablation Study on feature fusion module at last stage. Classification experiments are done on ModelNet40 without data augmentation.}
    \resizebox{0.7\columnwidth}{!}{
    \begin{tabular}{c|ccc}
        \hline
        \text{Method} & $z/z$ & $z$\text{/SO(3)} & \text{SO(3)/SO(3)} \\
        \hline
        \text{Max pooling} & $87.4$ & $87.4$ & $87.4$ \\
        \text{Average pooling} & $87.6$ & $87.6$ & $87.6$ \\
        \text{Pooling before fusion} & $86.6$ & $86.6$ & $86.6$ \\
        \text{Fusion before pooling} & $\textbf{88.3}$ & $\textbf{88.3}$ & $\textbf{88.3}$ \\
        \hline
    \end{tabular}
    }
    \label{ablation}
\end{table}

\section{Conclusion}
In this paper, we take a thorough study on PCA based rotation invariant deep point cloud analysis, and propose a universal framework which can be combined with common backbones. Various experiments show the effectiveness of our method, as the performance of our method is close to the raw backbone on aligned datasets while keeping consistent under any rotation.

\bibliographystyle{IEEEbib}
\bibliography{strings,refs}

\end{sloppypar}
\end{document}